\documentclass[a4paper, 10pt, conference]{ieeeconf}      

\IEEEoverridecommandlockouts                              

\usepackage{amsmath,amssymb,amsfonts}
\usepackage{graphicx}
\usepackage{amstext} 
\usepackage{array}   
\usepackage{makecell}
\usepackage{booktabs}
\newcolumntype{L}{>{$}l<{$}}
\DeclareGraphicsExtensions{.pdf,.png,.jpg,.jpeg}
\graphicspath{{figures/}{pictures/}{images/}{./}} 

\title{\LARGE \bf
Simple yet efficient real-time pose-based action recognition
}

\author{Dennis Ludl, Thomas Gulde and Crist\'obal Curio\\
	\scriptsize \\
\thanks{D. Ludl, T. Gulde and C. Curio are with the Cognitive Systems Group, Computer Science Department, Reutlingen University, Germany.
{\tt\small \{Dennis.Ludl, Thomas.Gulde, Cristobal.Curio\}@Reutlingen-University.de}
}}

\begin{document}

\begin{minipage}{\textwidth}\ \\[14pt] \centering
  \copyright 2019 IEEE. Personal use of this material is permitted.  Permission from IEEE must be obtained for all other uses, in any current or future media, including reprinting/republishing this material for advertising or promotional purposes, creating new collective works, for resale or redistribution to servers or lists, or reuse of any copyrighted component of this work in other works.
\end{minipage}

\newpage

\maketitle

\thispagestyle{empty}
\pagestyle{empty}

\begin{abstract}
Recognizing human actions is a core challenge for autonomous systems as they directly share the same space with humans. Systems must be able to recognize and assess human actions in real-time. In order to train corresponding data-driven algorithms, a significant amount of annotated training data is required. We demonstrated a pipeline to detect humans, estimate their pose, track them over time and recognize their actions in real-time with standard monocular camera sensors. For action recognition, we encode the human pose into a new data format called Encoded Human Pose Image (EHPI) that can then be classified using standard methods from the computer vision community. With this simple procedure we achieve competitive state-of-the-art performance in pose-based action detection and can ensure real-time performance. In addition, we show a use case in the context of autonomous driving to demonstrate how such a system can be trained to recognize human actions using simulation data.
\end{abstract}

\section{Introduction}
There is an increasing consensus that a human-like understanding of human behavior is a major challenge for autonomous systems, like self-driving cars in urban areas \cite{brooksBigProblemSelfDriving2017}. In the future, autonomous systems and human beings will co-exist in shared public spaces. Reliably inferring the world state with series sensor technology is still a challenge. One area that we consider very important is the detection of human actions. This area is still an open field and there are no systems that can be used productively in a stable and reliable manner. In areas where autonomous systems have to interact with people, it is very important that they have information about what people are exactly doing in their immediate environment. This is especially true if a direct interaction with the human being is to take place. Since human actions are highly dynamic, it is not only important to predict the actions correctly but also in real-time.

In addition to the runtime requirement for an algorithm, data-driven algorithms require massive amounts of training data. Data acquisition is usually one of the main problems in the development of a data-driven algorithm, thus we consider the provision of sufficient data to an algorithm to be an important factor in the design of this algorithm. We have shown in \cite{ludlUsingSimulationImprove2018} that we can train pose recognition algorithms with simulated data to recognize corner cases. As a continuation of this work we see great potential in the application of simulated data for the training of action detection algorithms. Since there is a domain shift from simulated visual data to real data, we decided to design a pose-based action recognition algorithm that works without direct dependence on visual sensor information. With this abstraction layer we want to enable the training of such an algorithm with simulated data and overcome domain transfer issues. This would save a lot manual effort that is required when recording and annotating real sensor data. 

\begin{figure}
  \centering
  \includegraphics[width=\columnwidth]{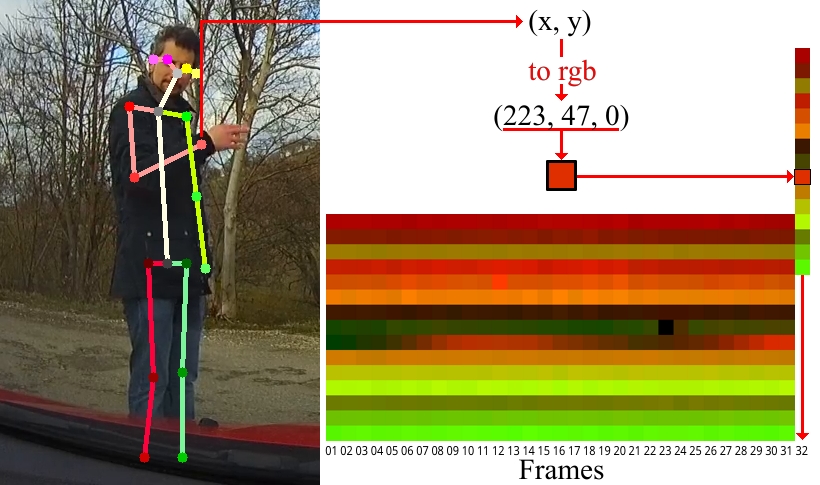}
  \caption{From skeletal joints to an Encoded Human Pose Image (EHPI), exemplified by the right wrist. The $x$ and $y$ coordinates are normalized (for the visualization the normalization process is simplified and ranges between 0 and 255 for RGB values), afterwards the X value is used as red component and the $y$ value as green component. This RGB value is set for each frame in an n-dimensional vector at a fixed location. In the example 15 joints are used, the right wrist is set in row 9. The full EHPI is of size $32\times 15\times 3$.}
  \label{fig:ehpi_process}
\end{figure}

Our current project \textit{Open Fusion Platform}\footnote{http://www.ofp-projekt.de/ (Last visited on 2019-04-08)} is about an autonomous vehicle with a valet parking function. It should automatically search for a free parking space on a parking lot and automatically be able to drive back to a pick-up point. Pedestrians can be present on the parking lot, thus it is important to recognize them. In addition to the pure recognition of pedestrians, it is also important to recognize what they are doing. In our use case, they are allowed to be in front of our parked vehicle, as long as the vehicle is not moving. In order to drive off while a pedestrian is detected in front of the vehicle, the pedestrian must clearly indicate, by a waving gesture, that the vehicle is allowed to drive out. Thus, the pedestrians have to be detected and further their current actions have to be classified. For this parking lot use case we have specified that the actions \textit{idle}, \textit{walk} and \textit{wave} must be detected.

Our contributions in this work are:

\begin{enumerate}
\item A recognition pipeline which operates on 2D monocular camera images in real-time. It contains functionality to detect objects, humans and their poses as well as to track and estimate humans and their actions.
\item A novel pose-based action recognition algorithm with state-of-the-art performance.
\item A demonstration on how to improve our action recognition algorithm with simulated data.
\end{enumerate}

\section{Related work}
There are various directions of research in the area of human action recognition. Some approaches are based on Convolutional Neural Networks (CNNs). They usually follow a multistream approach \cite{huangHumanActionRecognition2019, tuMultistreamCNNLearning2018, zolfaghariChainedMultistreamNetworks2017}, which uses an RGB image for visual feature extraction as well as a representation of the temporal flow, usually in the form of optical flow. There is also work which make use of human poses, either using pose directly \cite{huangHumanActionRecognition2019, zolfaghariChainedMultistreamNetworks2017} or apply some attention like mechanism to get visual features from important areas around the human skeleton \cite{tuMultistreamCNNLearning2018, duRPANEndtoEndRecurrent2017}. Those approaches often rely on recurrent neural networks \cite{angeliniActionXPoseNovel2D2018, duRPANEndtoEndRecurrent2017, songEndtoEndSpatioTemporalAttention2017}. Other approaches rely on handcrafted features extracted from human pose \cite{fangOnBoardDetectionPedestrian2017, garbadeHandcraftingVsDeep2016}. Most similar to our work is the work of Choutas et al. \cite{choutasPoTionPoseMoTion2018}. They encoded time information in human body joint proposal heatmaps with color and use this stacked, colored joint heatmaps as an input for a CNN to classify the action. To reach state-of-the-art performance they combined this approach with another multistream approach~\cite{carreiraQuoVadisAction2017}. Most of these approaches are relatively complex and therefore do not meet the real-time requirements of autonomous systems. Our approach is much simpler and still delivers competitive performance.

Not less important than action recognition algorithms is the generation of action recognition datasets. Khodabandeh et al. \cite{khodabandehDIYHumanAction2018} provide a method to automatically generate an action recognition dataset by partitioning a video into action, subject and context. Souza et al. \cite{desouzaProceduralGenerationVideos2017} proposed a database of simulated human actions. They used motion capture data containing action annotations combined with 3D human models in a simulated environment and show that it improves action recognition rates when combined with a small amount of annotated real world data. Other simulations containing animated humans exists \cite{muellerSim4CVPhotoRealisticSimulator2018, dosovitskiyCARLAOpenUrban2017}, but does not have a strong focus on realistic human actions. Nevertheless, the works in simulations to train and evaluate algorithms show that there is a large demand of realistic human motion data. Most current work either contains basic representations of humans or represents bigger databases which were generated procedurally targeting a broad range of motions. Human actions, on the other hand, must be simulated very precisely in order to be significant, especially when it comes to interactions. We have shown in previous work, that it is possible to easily finetune neural networks for pose detection by simulation in such a way that they can detect corner case poses, where they did not deliver any or insufficient results before \cite{ludlUsingSimulationImprove2018}. Following this approach we demonstrate how motion capture driven simulation is a useful method to generate human action recognition training data.

A drawback of simulated training data is the transfer of algorithms trained on simulated data to the application on real world data. There is usually a domain shift, which should be minimized by domain adaptation algorithms. Such approaches include the common use of few real data and many simulated data \cite{vazquezVirtualRealWorld2014}, methods which use decorrelated features \cite{sunVirtualRealityFast2014} and more advanced domain confusion algorithms \cite{tzengAdversarialDiscriminativeDomain2017}. As an open field of research it is important to find alternatives to avoid the domain transfer problem. We view the abstraction of input data as a promising approach to apply algorithms trained on simulated data directly to real data.

\section{Recognition Pipeline}
\label{sec:recognition_pipeline}
\begin{figure*}
  \centering
  \includegraphics[width=\linewidth]{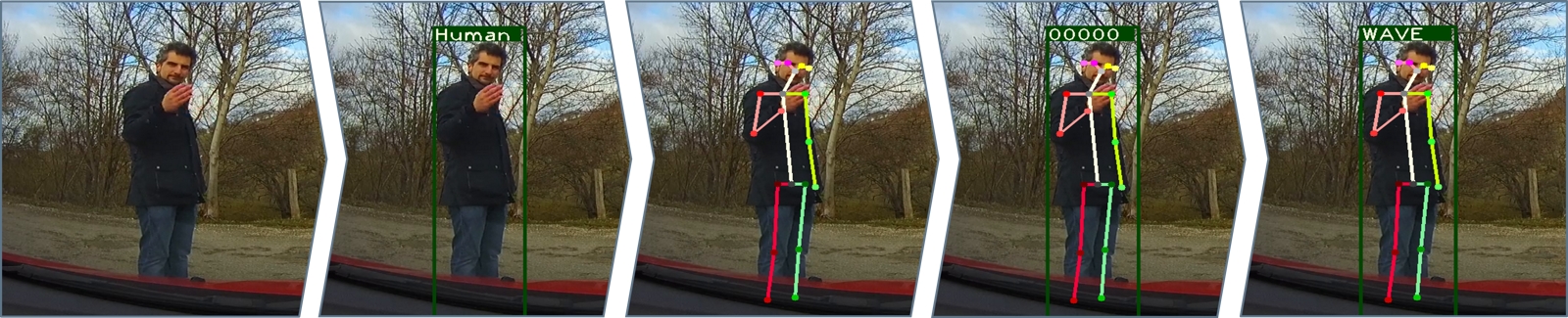}
  \caption{Real-time action recognition pipeline from a monocular camera mounted in the car observing the gesture of a potential user of the autonomous vehicle. From left to right: 1) Raw camera image, 2) Object detection~\cite{redmonYOLOv3IncrementalImprovement2018}, 3) Pose Recognition~\cite{xiaoSimpleBaselinesHuman2018}, 4) Pose-based human tracking and 5) Pose-based action recognition}
  \label{fig:teaser}
\end{figure*}
All steps in our pipeline\footnote{Code available at https://github.com/noboevbo/ehpi\_action\_recognition} are shown in Figure \ref{fig:teaser}. Each step is described in more detail in the following subsections. The action detection is described in a separate section. In general each step in this pipeline is based on the output data from the previous step. In the first step, we recognize objects in a 2D camera image, especially people (c.f. section \ref{sec:object_detection}). From this detection step we get bounding boxes around the respective object. If a person is recognized in a frame, we apply a pose recognition algorithm to the image information within that person's bounding box (c.f. section \ref{sec:pose_estimation}). The pose recognition is a single person solution and has to be done for every human in the image. Based on the human poses in successive frames we developed a pose-based tracking (c.f. section \ref{sec:human_tracking}) of humans. Based on the tracked human poses we finally perform an action detection (c.f. section \ref{sec:pipeline_action_rec}).

We decided to not use an end-to-end approach from sensor input to action detection, but to use a modular pipeline from object detection to actual action detection. As can be seen in the pipeline (c.f. Figure \ref{fig:teaser}), only object detection is performed on the entire sensor image. All other algorithms work on image regions and further pure pose data. This allows the more complex algorithms, such as pose recognition, to be applied only specifically, e.g. if a person is close to the autonomous system, or in the case of autonomous vehicles, if a construction worker has been identified. This also gives us the opportunity to use different training data at different steps and with different levels of abstraction layers, which enables us in particular to train our action recognition algorithm with simulated data (c.f. section \ref{sec:train_simulated}).

\subsection{Object Detection}
\label{sec:object_detection}
An object detection algorithm is used to obtain an initial estimate of human's presence in the image. In addition to the accuracy of the algorithm, the running time is the main criterion for the selection of the object detection algorithm. Possible false detections of the object detection algorithm can be compensated by the pose detection and the tracking of humans (c.f. section \ref{sec:human_tracking}). In this paper we use Yolo V3 \cite{redmonYOLOv3IncrementalImprovement2018} as a compromise between runtime and accuracy, which was pre-trained on ImageNet~\cite{dengImageNetLargescaleHierarchical2009} and the MSCOCO~\cite{linMicrosoftCOCOCommon2014} dataset. The object detection algorithm can be replaced by alternative object detection algorithms, depending on accuracy and runtime requirements. The input into the object detector is an RGB camera image, which may be scaled down to allow faster processing. The algorithm then estimates possible object locations in the image in the form of bounding boxes and classifies their content. After postprocessing the resulting data is a list of classified bounding boxes. In this work, only bounding boxes that have been classified as humans, are used.

\subsection{Pose Estimation}
\label{sec:pose_estimation}
For the human pose estimation we use the approach from Xiao et al.~\cite{xiaoSimpleBaselinesHuman2018} with its network pre-trained on the MSCOCO~\cite{linMicrosoftCOCOCommon2014} and MPII~\cite{andriluka2DHumanPose2014} datasets. The algorithm requires human bounding boxes as input and estimates a human skeleton in this cropped region. Like in most state-of-the-art pose recognition algorithms, a heat map is predicted for each joint, indicating the estimated probability for each joint. During the post-processing non maximum suppression is performed and a human skeleton is reconstructed in the form of 2D joint positions and their connections.

\subsection{Human Tracking}
\label{sec:human_tracking}
For our action recognition algorithm, a person's skeletal information is required across multiple frames. Since the pose recognition algorithm described above is applied to single images, the skeletons in several frames are initially independent of each other. In order to establish the reference of skeletons across several frames, we track the skeletons based on their joint positions. We use the pyramidal implementation of the Lukas Kanade Feature Tracker \cite{bouguetPyramidalImplementationLucas2000} and take the joint positions of the human skeletons in the image as features to be tracked. We end up with an estimated skeleton in frame $n$ for each skeleton in frame $n-1$. With these tracked skeletons from frame $n-1$ and new detected skeletons from frame $n$ we have a number of skeleton proposals which need to be merged as follows. 

A merge is done by measuring the similarity of two human poses. If they are similar enough the two skeletons will be merged to one human skeleton, thus tracking the human over time. In addition to comparing detected and tracked people for a possible merge, all detected people must also be compared, since the same person could be detected several times by false detection of the object or pose recognition algorithm.

The first step to merge two humans is to find the similarity between two human skeletons. Let $a$ and $b$ be human skeleton hypotheses from a list of detected or tracked skeletons. We define $\Delta_{a}$ as the maximum distance between joint $i$ in two skeletons to be considered being part of the same skeleton. $\Delta_{a}$ is calculated by using the bounding box width $w_a$ and height $h_a$ of human $a$ (c.f. Equation \ref{eq:delta_a}).

\begin{equation}
  \label{eq:delta_a}
  \Delta_{a} = F(\sqrt{w_{a}^2 + h_{a}^2})
\end{equation}

Factor $F$ denotes a hyperparameter, which corresponds to the percentage of the human's bounding box diagonal. Setting $F=0.025$ has proven in practice. Next, the Euclidean distance $\delta_{ab_{i}}$ of joint $i$ between human skeleton $a$ and $b$ is calculated (c.f. Equation \ref{eq:joint_distance}). 

\begin{equation}
  \label{eq:joint_distance}
    \delta_{ab_{i}} = \rVert a_i - b_i \rVert_2
\end{equation}

This distance is only considered if $a$ and $b$ contain joint $i$ with a minimum probability of $T_J$ that specifies the minimum joint quality required to use joints in the tracking process. Setting $T_J=0.4$ worked well in practice. This constraint is included to enable tracking even when some joints are not recognized or only poorly recognized, e.g. due to occlusion. 
We then calculate the similarity score ($S_{ab_{i}}$) for joint $i$ by comparing the actual joint distance with the maximum acceptable distance (c.f. Equation \ref{eq:joint_similarity}).

\begin{equation}
  \label{eq:joint_similarity}
    S_{ab_{i}} = \begin{cases}
        1 - (\delta_{ab_{i}} / \Delta_{a}),& \text{if } \delta_{ab_{i}} < \Delta_{a}\\
        0,              & \text{otherwise}
    \end{cases}
\end{equation}

The similarity ($S_{ab}$) between human skeleton $a$ and $b$ is calculated by combining all joint similarities $S_{ab_{i}}$ (c.f. section \ref{eq:skeleton_similarity}). 

\begin{equation}
  \label{eq:skeleton_similarity}
    S_{ab} = \frac{1}{I}\sum\limits_{i=1}^I (s_{ab_{i}})
\end{equation}

Factor $I$ denotes the number of joints used for tracking. We then try to merge all detected human skeletons with other detected human skeletons to avoid repeated recognition of skeletons belonging to the same person. We define a threshold $T_S=0.15$ to specify when two human skeletons are similar. If the similarity score is above $T_S$ for two detected humans the detection with the lower score is removed from the detection list. After merging the detected human skeletons they are merged with all tracked humans from previous frames. In this merge process every detected human is compared to every tracked human. If the similarity score of two skeletons is above $T_S$ the identifier of the tracked human is assigned to the detected human and the tracked skeleton is removed from the tracking list. If after the merge process some tracked humans are left over, meaning that the object detector did not provide a human proposal at the location of a tracked human, we apply the pose estimation on the bounding box of the tracked human and if a human skeleton with a score higher than $T_J$ is estimated we keep this human with its identifier as a detected human. With this approach we can compensate false negatives from the object detector and we are even able to deactivate the object detection completely to improve performance (c.f. section \ref{sec:performance}).

\subsection{Performance}
\label{sec:performance}
The runtime of our pipeline scales with the number of people to be detected. Usually only the humans in the immediate surrounding area of the autonomous system are relevant, so the entire pipeline may normally not have to be used for all humans in the image. For one human with input image resolution of 1280x720, downscaled for processing to 640x360, the entire pipeline runs on average with 29 FPS. For two people, the FPS is reduced to around 21 FPS, which still ensures real-time processing. To improve performance in special cases, object detection, which is usually performed for each frame, can be disabled once a person has been detected. The pose and action detection can then be continued for this person based on bounding box proposals from our tracking process. By switching off the object detection, on average 57 FPS can be achieved for one person. Depending on the requirements, it would be possible to perform object detection only on a limited number of frames. It is also important to note that our implementation is not completely designed for performance, as more emphasis was placed on code readability. It can be assumed that the performance can be increased with appropriate adaptations. All performance tests were carried out on a laptop with an Intel i7-8700 six core CPU and a NVIDIA GTX 1080 GPU using Ubuntu 18.04 with CUDA 10.0 and CUDNN 7.4.2.

\section{Pose-based action recognition}
\label{sec:pipeline_action_rec}
  
Since a lot of progress has been made in the field of convolutional neural networks, we have decided to investigate an approach in which human skeletons are encoded over time in an image-like data structure. We expected a more stable and accurate system compared to recurrent neural networks. 

The basic process is shown above in Figure \ref{fig:ehpi_process}. Once the pose of a human has been extracted from the camera image, the basic idea is to encode the $x$, $y$ and $z$ positions of the joints as red, green and blue values in an RGB image. In this paper we work with 2D pose detection on monocular camera images, so the $z$ value is not used and thus the blue channel is set to zero. The channel could also be removed as long as only the $x$ and $y$ coordinates are used. In order to convert the global joint coordinates into corresponding 'color values' we normalize them. This process is described in more detail in section \ref{subsec:preprocessing}. Note that we normalize the values as network input in the continuous range from zero to one and not as discrete integer values from $0-255$ thus the analogy of an image is therefore not entirely accurate. Yet for visualization purpose in this paper we normalized the joint positions to $0-255$ for the figures. Basically any number of joints can be encoded. In the current work we use nose, neck, hip center, left shoulder, left elbow, left wrist, right shoulder, right elbow, right wrist, left hip, left knee, left ankle, right hip, right knee and right ankle in this particular order. The encoded joints are assigned in a fixed order in a $1\times n\times3$ matrix, where $n$ stands for the number of joints. After the human pose for a frame has been encoded into such a matrix, it is appended as last column to a $m\times n\times3$ matrix and replaces the first column of this matrix if it already contains $m$ frames. Each column represents an encoded human pose in a frame. The full matrix represents an Encoded Human Pose Image. In the current work we use $m=32$ which were chosen because we want to analyze about one to two seconds of movements to give an action estimate. Additionally, it corresponds to standard image width used in machine learning applications. To stabilize our action recognition we take the recognitions of the last $20$ frames and use the action class with the highest summed probability in the last $20$ frames as our prediction.

\subsection{Network}
\begin{figure}
  \centering
  \includegraphics[width=\columnwidth]{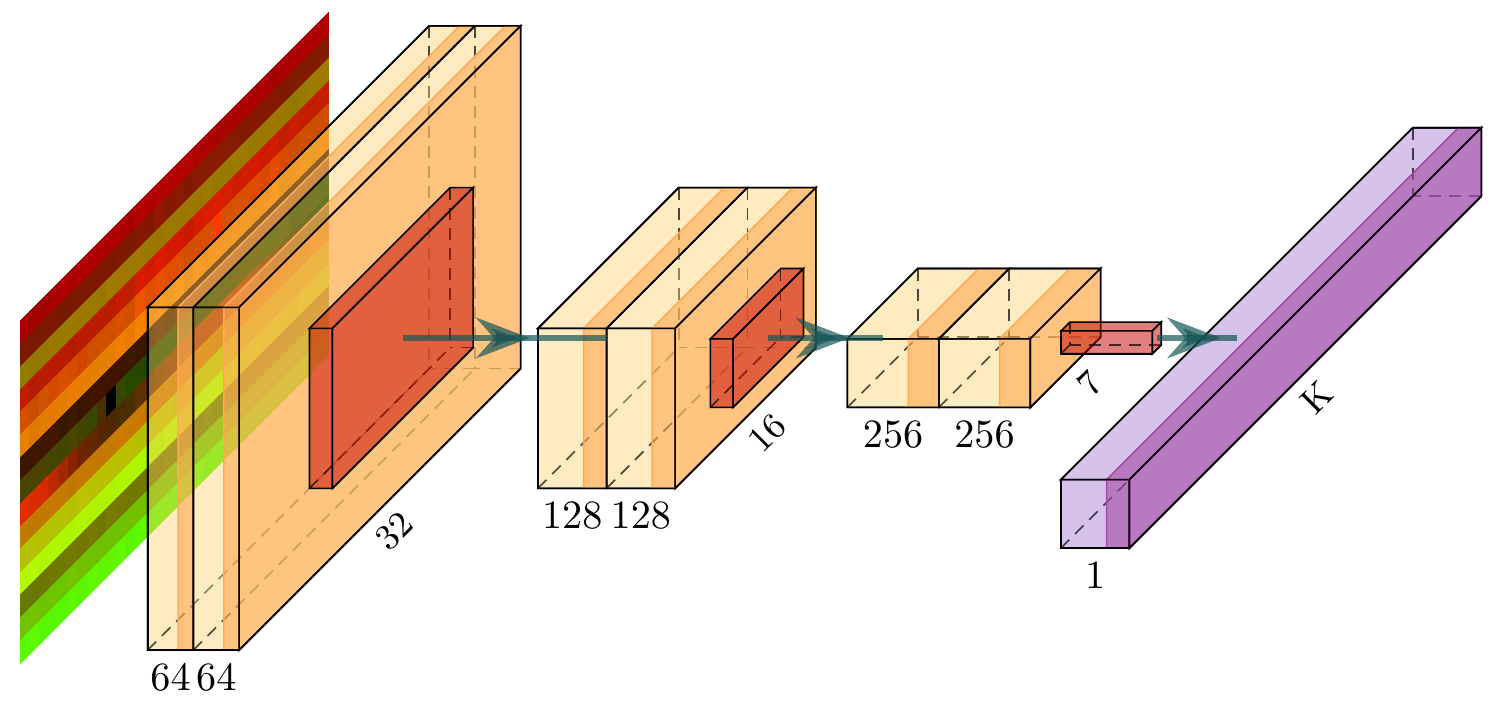}
  \caption{Simple network architecture that we used to classify the EHPIs on the JHMDB dataset.}
  \label{fig:ehpi_jhmdb_network}
\end{figure}
For the classification of the EHPIs we used a very simple network consisting of six convolutional layers, of which each has a $3\times 3$ kernel as well as both a padding and stride of one. A fully connected layer is placed at the end for the final action classification (see Figure \ref{fig:ehpi_jhmdb_network}). Each convolutional layer is followed by batch normalization~\cite{ioffeBatchNormalizationAccelerating2015}. As activation function we use ReLU in the convolutional layers. After the second and fourth convolution we apply a max pooling layer with a kernel size of $2\times 2$ which reduces the spatial resolution by factor two. After the last convolutional layer we apply a global average pooling layer. We use Xavier initialization \cite{glorotUnderstandingDifficultyTraining2010} for all convolutional layers. By using such a deep neural network architecture the deeper the network is, the more spatio-temporal context should be encoded in the learned features due larger receptive fields.

Since we have used considerably more data in our use case than is available in the JHMDB~\cite{jhuangUnderstandingActionRecognition2013} dataset (see section \ref{sec:datasets}), the network is no longer sufficient. Expanding the network with further convolutional layers and also increasing the size of the fully connected layer would result in the network having more parameters than some existing and efficient CNNs for classification. Therefore we employ the ShuffleNet v2~\cite{maShuffleNetV2Practical2018} architecture with which we also demonstrate the application of standard computer vision algorithms to EHPIs.

\subsection{Preprocessing}
\label{subsec:preprocessing}
The normalization of the EHPI takes place on the entire $m\times n\times3$ matrix. We normalize the encoded $x$ and $y$ values independently between zero and one. This type of normalization is intended to ensure the independence of the body size of different people while maintaining the relative change in scale through a different distance to the camera. We consider correspondingly the local range of motion of a person for a time window of length $m$. Before normalization, we also remove human body joints that are outside the image as a preprocessing step by setting their coordinates to zero. When joints are not recognized or have a probability below $T_J$ the $x$ and $y$ values for them are set to zero. The same applies for human poses which are not recognized at all, here we set the complete $1\times n\times3$ matrix to zero. We define that an EHPI requires at least two frames with human poses to be considered.

\subsection{Examples}
\begin{figure}
  \centering
  \includegraphics[width=\columnwidth]{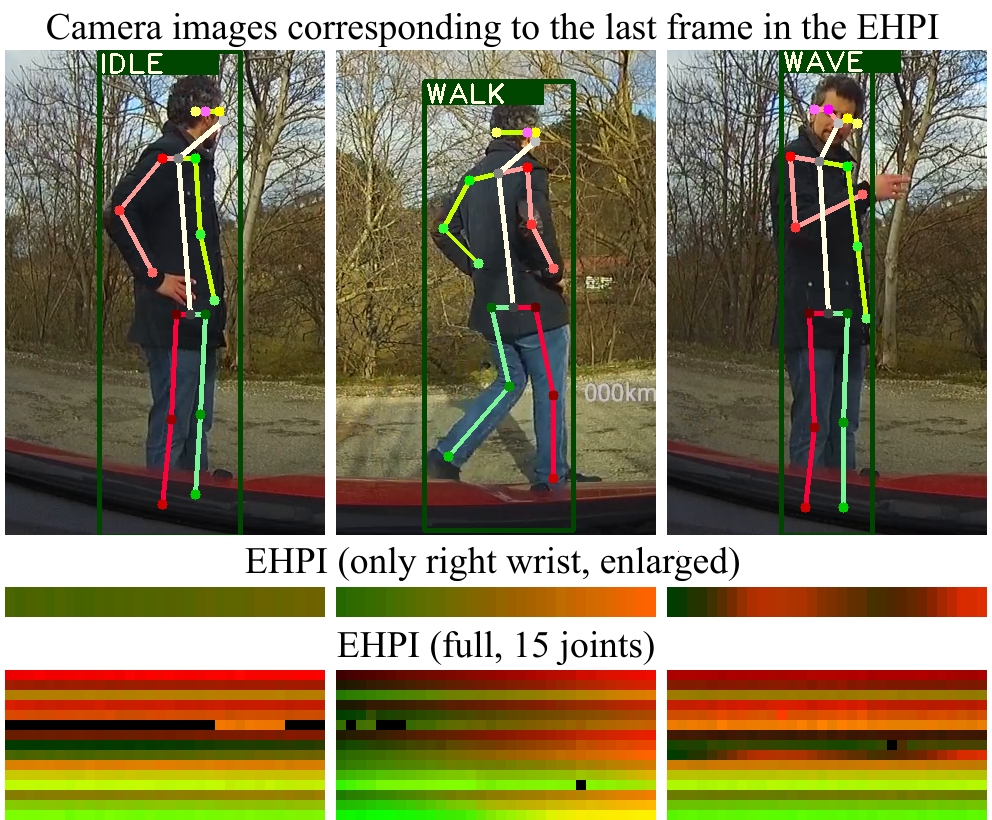}
  \caption{EHPI examples of different actions. The example of the right wrist, which is explicitly shown at three times its height, clearly shows that a smooth color gradient is visible in the idle action, a color gradient from green to orange is visible during walking and a repetitive gradient from green to red is observable during waving.}
  \label{fig:ehpi_example}
\end{figure}
Figure \ref{fig:ehpi_example} shows the EHPIs and a camera image of the last frame (rightmost column) of the EHPI for three examples of the actions \textit{idle}, \textit{walk} and \textit{wave}. The row that represents the joint of the right wrist is plotted in an enlarged view since it is diagnostic for discriminating the actions of interest in the following example. For the action \textit{idle} the color representation is relatively constant over the whole period, because there is hardly any movement of the joint. For the action \textit{walk}, one can notice a smooth transition from green to orange. This is due to the fact that the joint of the right wrist moves from left to right of the image during the EHPI period. Therefore the normalized $x$ value moves more and more towards one (in the visualization thus the red value towards 255), while the $y$ value (in the visualization the green value) remains relatively constant. During the \textit{wave} action one can notice a repetitive color gradient from green to red, because the joint of the right hand moves repeatedly in $x$ direction during the wave movement.

\begin{figure}
  \centering
  \includegraphics[width=\columnwidth]{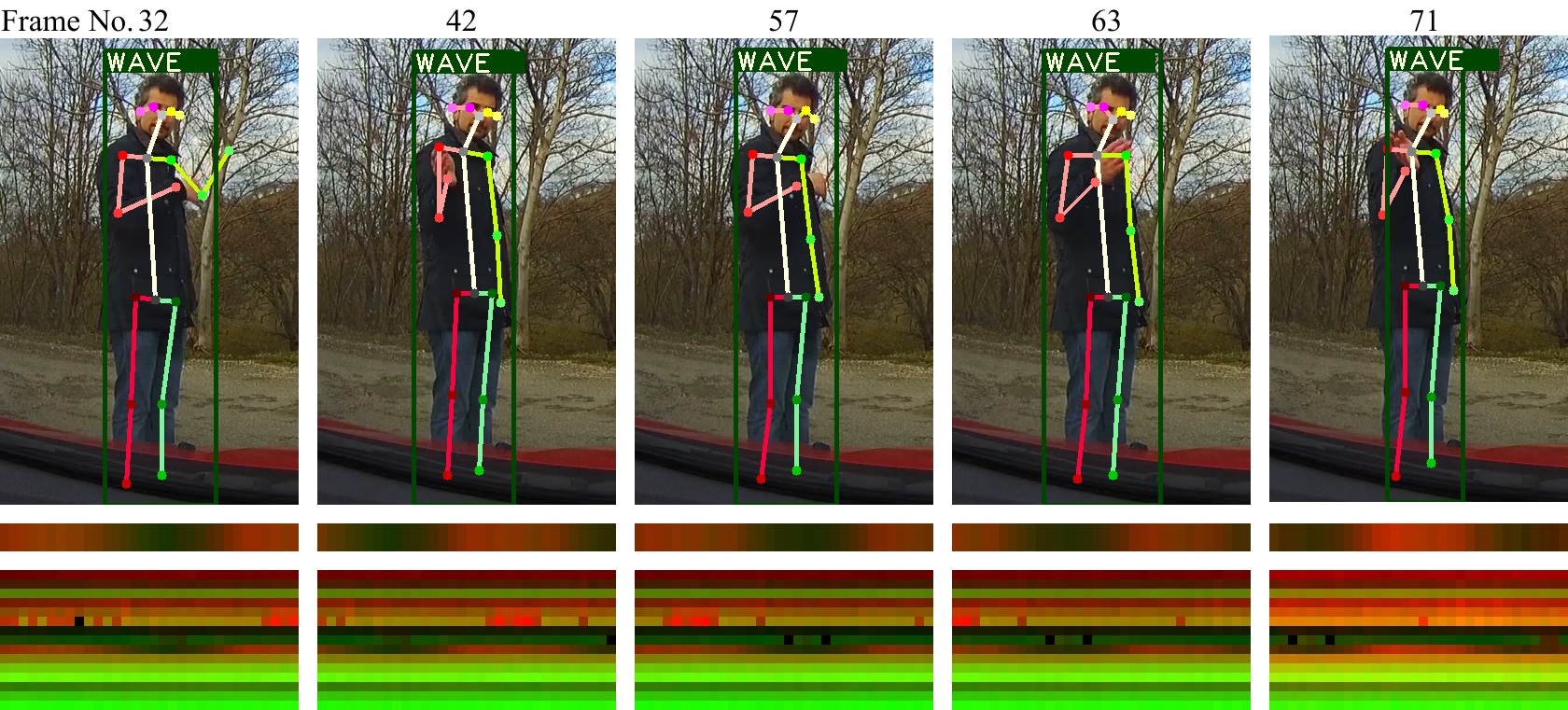}
  \caption{Five frames from a sequence with camera image, EHPI (right wrist, enlarged) and the full EHPI. The EHPI for the right wrist moves during waving towards red (max. in $x$ direction). The first picture shows a false detection of the left wrist (EHPI, row 6), which is filtered by the application of also noisy training data of the action detection. In the last image the whole EHPI is shifted more into red. This is due to the fact that there are no more extreme false detections of the left wrist that shift the maximum $x$ value during normalization.}
  \label{fig:example_sequence}
\end{figure}

Figure \ref{fig:example_sequence} shows a sequence of a wave movement in more detail. At the end (right) part of the EHPI for the joint of the right wrist, the color encoding gets more red when the joint is on the right side of the picture. Further, the effects of false detection of body joints is displayed clearly. In this example, the left ankle (row 6 in EHPI) is partially recognized incorrectly and thus is encoded much further to the right than it actually is. This becomes clear with the strongly red coded areas, which appear without a clean transition from green to red. The joint of the left ankle was recognized with a probability above $T_J$ by the pose estimation algorithm, thus it is not encoded as zero. Due to the distorted, more extreme red values of the left hand, the entire EHPI is shifted a little into the green area, which becomes clear in the last image (frame 71), where most of the joint recognition errors are no longer present and the color of the entire EHPI shifts towards red.

\section{Datasets}
\label{sec:datasets}
The \textbf{JHMDB}~\cite{jhuangUnderstandingActionRecognition2013} dataset consists of 928 videos of which each has an annotation label denoting one of 21 action classes. Each video has a resolution of 320x240 pixels. The evaluation is done on three splits of which each uses about 30\% test data. Results are reported as mean over all splits. From here on JHMDB refers to the full dataset, while JHMDB-1 refers to JHMDB split 1 with pose data from our pipeline and JHMDB-1-GT refers to JHMDB-1 with pose data from the JHMDB ground truth.

Our automotive parking lot use case dataset \textbf{SIM} consists of various camera sequences. We have recorded different videos with a Logitech C920 webcam, an iTracker GS6000 dashcam and a Yi 4k+ camera. It is a very use case specific dataset, which only contains the actions \textit{idle}, \textit{walk} and \textit{wave}. Recordings were partly taken inside buildings, partly also in use case situations in the vehicle. In addition, our dataset contains some simulated elements, which are described in more detail in section \ref{sec:train_simulated}. The entire dataset consists of 216 labeled sequences and a total of 61826 EHPIs. All videos have a resolution of 1280x720 at 30 FPS. All sequences contain actions from the same person. For the evaluation 27 sequences with a total of 8351 frames are used. All sequences are cuts from one scene, which corresponds to our use case. The scene was recorded simultaneously from the dashcam and the action cam to get data from two different sensors on slightly different locations.

\subsection{Simulated data}
\label{sec:train_simulated}
\begin{figure}
  \centering
  \includegraphics[width=\columnwidth]{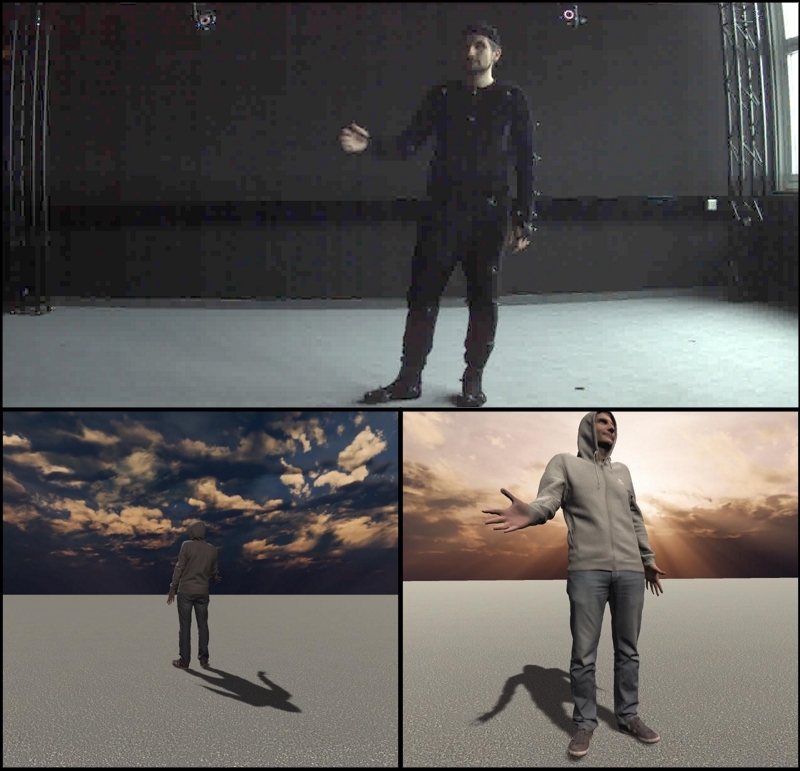}
  \caption{Demonstration of virtual sensor information used to train our action recognition algorithm. On top a picture of a real sensor of the motion recording in the motion capture lab and below the simulated scene from two different camera positions.}
  \label{fig:training_data}
\end{figure}

Due to our positive experiences with the use of simulation to improve pose recognition algorithms \cite{ludlUsingSimulationImprove2018}, we decided to use simulation data in this work to further enrich our training data. The advantage of our modular pipeline is that we can use simulation data as training data at different steps in the pipeline, while real data is used at other steps. In the case of action detection, this offers the great advantage that we have the abstraction layer of the pose data between the sensor information and the action detection. The underlying hypothesis is that sensor domain transfer problems between simulation and real data are prevented by this abstraction layer. Motion data is required to generate the simulation data. In principle, the motion capture data alone is sufficient to generate ground truth data for our pose-based action detection. By a corresponding 2D projection of the 3D joint coordinates for any number of camera positions in 3D space, 2D pose information can be obtained without generating camera sensor simulations. Since pose recognition algorithms are not perfect and artifacts like a slight jittering of the joint positions, false recognition of joints or not recognizing joints can occur, we additionally generate the simulated camera images to apply the pose recognition algorithms and use the output as ground truth with such kind of natural noise for the action recognition. In previous work we have shown evidence that pose recognition algorithms have similar problems on simulated data as on real data~\cite{ludlUsingSimulationImprove2018}, thus we expect to see the same bias on estimated human poses on simulation data as on real data. To simulate sensor information it is also necessary to use a 3D environment and a 3D human model. We recorded the motion data of the actions \textit{idle}, \textit{walk} and \textit{wave} in our motion capture laboratory. One person performed every action ten times for ten seconds. The motion data is used to animate a 3D human model in our Unity\textregistered~based simulation. The environment is not very relevant in this case, because we only need to recognize an animated 3D human model with our pose recognition algorithm and using its output as ground truth for action recognition. Therefore we only use a flat area with a skybox for the background without any other environmental details (c.f. Figure \ref{fig:training_data}). At the end, various virtual camera sensors can be placed around the person, which then generate corresponding sensor information. For each virtual camera image the ground truth, in this case the 2D pose and the corresponding action, can be generated automatically \cite{ludlUsingSimulationImprove2018}. Figure \ref{fig:training_data} shows some examples of the simulated sensor information. We used a total of six camera positions in this work.

In the following we distinguish between SIM (gt) which contains the perfect pose data from motion capturing directly, SIM (pose) which contains the pose data from the output of our pose detection pipeline and SIM which contains the data from both sources.

\subsection{Data Augmentation}
To increase the variance in our training data we use data augmentation. Joints are flipped horizontally in 50\% of the cases. If the image is flipped, in 50\% of the flipped images the indexes of the left and right joints are also switched, thus we simulate in 25\% of the cases a person looking in the other camera direction than originally. In addition, we partly remove joints to simulate occlusion. In 25\% of the cases we remove the joints of both feet and in 6.25\% of the cases we also remove the joints of the knees. This type of augmentation is mainly the result of our use case, in which it can happen that the feet and partly also the knees are covered by the hood of the vehicle.


\section{Training}
We used 33\% of the JHMDB-1 training data for validation. We did not focus on the tuning of hyperparameters and therefore only varied the batch size, the learning rate and the number of epochs to find out how to train the network fast and stable. We trained the network with a batch size of 64, an initial learning rate of 0.05 and a standard momentum of 0.9 for 200 epochs (140 on JHMDB-1-GT) for our experiments on the JHMDB dataset. Since the JHMDB data set is quite small, we also used a weight decay (L2 regularization~\cite{ngFeatureSelectionL12004}) of $5e^{-4}$ to counteract overfitting. For optimization we use the stochastic gradient descent (SGD) and the cross-entropy loss. We reduce the learning rate every 50 epochs by a factor of ten. The classes in the JHMDB data set are not evenly distributed, especially as far as the number of EHPIs per video sequence is concerned, as they vary in length. Therefore, we apply a sampling per epoch that outputs a balanced number of samples for each class by reusing samples from classes with few samples and using only a subset of samples from classes with many samples. We use the same parameters for our use case data set, but since there is considerably more data available, we adjust the batch size to 128. We train the network with five different seeds to exclude random effects during weight initialization and to ensure the reproducibility of our results. We therefore report results as the mean value with standard deviation over these five runs.




\section{Experiments}
\subsection{JHMDB evaluation}
The current state-of-the-art approach PoTion \cite{choutasPoTionPoseMoTion2018} combines their pose-based action detection with the multi-stream approach I3D \cite{carreiraQuoVadisAction2017} and achieves a total performance of 85.5\% on the JHMDB data set. Since we only need the three actions \textit{idle}, \textit{walk} and \textit{wave} for our use case, which can also be distinguished purely with pose data, and real-time is a necessary prerequisite, we use the pure pose-based EHPIs in this work. We therefore compare ourselves with parts of other work that also report results for pure pose-based algorithms. The results are summarized in Table \ref{tab:jhmdb} in terms of accuracies. JHMDB results are reported as mean value over the three dataset splits. In cases where our pose recognition pipeline was unable to find a human in a video sequence we could not apply the action recognition algorithm and thus we counted that sample as recognized falsely. In 904 of 928 videos we were able to recognize a human skeleton in at least two frames and thus created an EHPI and performed the action detection. For cases where we detected more than one person in a video we used the one with the highest pose score.

\begin{table}[!htbp]
  \centering
    \caption{Results on the JHMDB dataset compared to other pure pose-based algorithms.}
  \begin{tabular}{llll}
    \toprule
  \thead{Method}                & \thead{JHMDB}                          & \thead{JHMDB-1} & \thead{JHMDB-1-GT}  \\
  \midrule
  PoTion\cite{choutasPoTionPoseMoTion2018}            & 57.0                                    & 59.1                    & \textbf{70.8} \\
  Zholfaghari et al.\cite{zolfaghariChainedMultistreamNetworks2017}    & N/A                                        & 45.5                    & 56.8 \\
  \textbf{EHPI (ours)}         & \textbf{60.5} $\pm$ \textbf{0.2}  & \textbf{60.3} $\pm$ \textbf{1.3}  & 65.5 $\pm$ 2.8\\
  \bottomrule
  \end{tabular}
    \label{tab:jhmdb} 
  \end{table}

On the whole JHMDB dataset we outperform PoTion~\cite{choutasPoTionPoseMoTion2018} by a margin of 3.5\%. On JHMDB-1 a result is also provided by Zolfaghari et al.~\cite{zolfaghariChainedMultistreamNetworks2017}. We outperformed PoTion by a margin of 1.2\% and Zolfaghari et al.~\cite{zolfaghariChainedMultistreamNetworks2017} by a margin of 14.8\%. Using only the ground truth pose information provided by the JHMDB Dataset the results of PoTion outperform our results by a margin of 5.3\%. This can be either caused because our pose recognition pipeline provides better pose information or because PoTion was applied to a cropped image around the actuator.

\subsection{Automotive parking lot use case evaluation}
To evaluate our system we compared two types of results. First, we show how many of the action sequences were correctly recognized, denoted by Accuracy (Seq). Since a sequence can sometimes last several seconds and the total detection consists of the accumulated predictions of the individual EHPIs, we also consider it useful to indicate how many of the individual EHPIs are correctly detected, denoted by Accuracy (EHPIs). The results are shown in Table \ref{tab:use-case}.

\begin{table}[!htbp]
  \centering
    \caption{Use Case Results}
  \begin{tabular}{l l l}
    \toprule
  \thead{Method}  & \thead{Accuracy (Seq)}  & \thead{Accuracy (EHPIs)} \\
  \midrule
  SIM (Pose)      & 80.74 $\pm$ 2.77          & 69.72 $\pm$ 1.80 \\
  SIM (GT)        & 79.26 $\pm$ 3.78          & 67.78 $\pm$ 1.86  \\
  SIM             & 81.48 $\pm$ 3.31          & 70.64 $\pm$ 2.60 \\
  Real only       & 99.26 $\pm$ 1.48          & 95.75 $\pm$ 1.65 \\
  SIM + Real      & 99.26 $\pm$ 1.48          & 97.07 $\pm$ 1.80 \\
  \bottomrule
  \end{tabular}
    \label{tab:use-case} 
  \end{table}

With real data only we are able to correctly classify 99.26\% of the test sequences. The misclassified sequence is an \textit{wave} sequence that has been classified as \textit{idle}. The false detection was probably caused by the fact that the waving in this sequence was executed with the left hand, for which only little training data was available. The overall great results are due to the fact that the use case is rather focused and we have enough similar training data available. With 81.48\% correctly recognized sequences and 70.64\% correctly recognized EHPIs when trained purely on simulated data, there seems to be no big domain shift between simulated and real training data. We also found that the performance is slightly better when we use the noisy pose data from our pose detection pipeline as ground truth rather than using the pose information directly from the motion capture system, hinting that it is beneficial to use both ground truth sources. As the standard deviation shows, the hyperparameters are not yet optimal for training, but in this paper the network tuning is not the focus. By combining real and simulated training data for action detection, we were able to increase the overall detection rate of all EHPIs by 1.32\% to 97.07\%. Considering how easily and quickly the simulated data can be generated, the use of the simulation approach, at least as an addition to real data, is very promising.

\section{Conclusion}
In this work, we have shown how an efficient pipeline can be built that can recognize humans in real-time, estimate and track their poses, and recognize their current action. We have introduced a new encoding technique, in which human poses are encoded over a fixed period of time into an image-like data structure that can be used for action recognition using classification CNNs. Our EHPI based action detection delivers state-of-the-art performance compared to other pose-based algorithms and still runs in real-time. In future work it should be investigate how further scene properties and context can be encoded into an EHPI in order to be able to recognize actions that are not distinguishable on pose data only. In addition, we were able to realize the requirements of our automotive parking lot use case with the presented pipeline. Action recognition results could be transferred to other sensors, environments and people in first tests. We were also able to show that the use of simulation data in combination with real data is suitable for the enrichment of training of action detection algorithms. The results obtained on the purely simulated training data are also very promising. This approach must be further evaluated to determine if the portion of real data for the training can be reduced further or even be omitted. Further, we have currently used standard image classification CNN architectures for the classification of the EHPIs. These do not take into account the special spatiotemporal structure of an EHPI. With more specific network architectures exploiting these spatiotemporal relationships between joints, the process could probably be further improved. 


\addtolength{\textheight}{-12cm}   

\section*{ACKNOWLEDGMENT}
This project has been sponsored by BMBF project \textit{OFP} (16EMO0114) and \textit{MoCap 4.0} (03FH015IN6).


\bibliography{itsc_2019_pb_action_rec.bib}
\bibliographystyle{IEEEtran}

\end{document}